%% file: main.tex
\documentclass[10pt,twocolumn,letterpaper]{article}
\usepackage[T1]{fontenc}
\usepackage{pgsd-preprint}
\usepackage{multirow,colortbl}
\usepackage{flushend}
\definecolor{studentrow}{HTML}{E4DCEB}
\definecolor{linkblue}{rgb}{0.21,0.49,0.74}
\usepackage[pagebackref,breaklinks,colorlinks,allcolors=linkblue]{hyperref}
\AtBeginDocument{\normalsize}
\title{Efficient World-Model Planning via Physics-Guided Search Distillation}
\newcommand{\authormark}[1]{\textsuperscript{#1}}
\author{%
  Hongyu Zhang\authormark{1,\dag},\enspace
  Xiaoshuai Hao\authormark{2,\dag},\enspace
  Heye Huang\authormark{3},\enspace
  Qingwen Meng\authormark{4},\enspace
  Wenjun Shi\authormark{1},\\[1pt]
  Jianao Xu\authormark{1},\enspace
  Xin Hu\authormark{5},\enspace
  Xuanru Jin\authormark{1},\enspace
  Jingyi Zhang\authormark{5,\ddag},\enspace
  Baicang Guo\authormark{1,*}\\[2pt]
  \small \authormark{1}Yanshan University,\enspace
  \small \authormark{2}Xiaomi Corporation,\enspace
  \small \authormark{3}KAIST,\enspace
  \small \authormark{4}Tsinghua University,\\[-1pt]
  \small \authormark{5}Beijing Automotive Technology Center Co., Ltd.\\[1pt]
  \footnotesize \authormark{\dag}Equal contribution.\quad
  \authormark{*}Corresponding author.\quad
  \authormark{\ddag}Project Leader.}
\hypersetup{pdftitle={\thetitle},
  pdfauthor={Hongyu Zhang, Xiaoshuai Hao, Heye Huang, Qingwen Meng, Wenjun Shi,
    Jianao Xu, Xin Hu, Xuanru Jin, Jingyi Zhang, Baicang Guo}}
\begin{document}
\maketitle
\input{sections/00_abstract}
\input{figures/figure1_layout}
\input{sections/01_introduction}
\input{sections/02_related_work}
\input{sections/03_method}
\input{sections/04_experiments}
\input{sections/05_conclusion}
\label{paper:last-text}
\clearpage
\raggedcolsend
\raggedbottom
\input{sections/06_llm_usage}

{\small
\bibliographystyle{appendix/ieeenat_fullname_citation}
\bibliography{references}
}
\clearpage
\flushcolsend
\input{appendix/appendix}

\end{document}

%% file: sections/00_abstract.tex
\begin{abstract}
\frenchspacing
World models support autonomous driving by evaluating candidate trajectories through predicted future scene evolution.
However, accurate future prediction does not ensure road-compliant execution; when all candidates violate road boundaries, re-ranking cannot select a feasible trajectory.
We propose Physics-Guided Search Distillation (PGSD), a single-view planning framework that combines future representations and execution feedback for verified beyond-pool repair through three components:
\textbf{(1) Execution-Aware Selection} checks ego-vehicle road compliance under simulated execution and selects the highest-scoring feasible trajectory;
\textbf{(2) a Clearance-Aware Search Teacher} generates beyond-pool corrections and selects demonstrations with larger minimum road-boundary clearance during execution to buffer imitation errors;
and \textbf{(3) a Future-Conditioned Repair Student} predicts one correction from rewarder features and execution feedback for single-rollout verification.
The pretrained planner and rewarder remain frozen; no future scenes are generated online.
On NAVSIM v1, single-view PGSD achieves 90.97 PDMS, exceeding all compared prior single-view planners and retaining 73.2\% of the search teacher's PDMS gain over execution-aware selection.
On NAVSIM v2 navhard, it reaches 38.96 EPDMS, improving on WorldDrive by 11.24 points (40.5\%).
The student replaces up to ten search rollouts with a single verification rollout per repair trigger.
HUGSIM Hard examples provide qualitative evidence of zero-shot transfer.
\end{abstract}

%% file: figures/figure1_layout.tex
\begin{figure}[!t]
\centering
\includegraphics[width=\columnwidth]{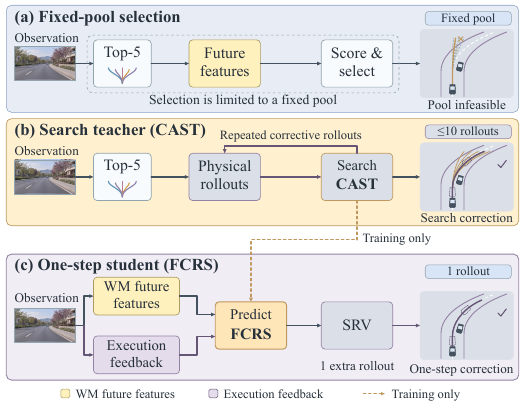}
\caption{\textbf{Existing selection and our search distillation.} (a) Existing future-aware scoring selects from a fixed candidate pool. (b) Our Clearance-Aware Search Teacher (CAST) generates beyond-pool corrections through repeated execution rollouts. (c) Our Future-Conditioned Repair Student (FCRS) combines world-model (WM) features and execution feedback, predicting one correction checked by Single-Rollout Verification (SRV). Rollout counts denote additional repair queries; dashed warm-gold arrows denote training-only teacher demonstrations.}
\label{fig:1}
\end{figure}

%% file: sections/01_introduction.tex
\section{Introduction}\label{sec:1}

End-to-end autonomous driving has attracted increasing attention by learning to map sensor observations to driving trajectories \cite{hu2023uniad,jiang2023vad,liao2025diffusiondrive}. Driving world models extend this capability by anticipating action-conditioned scene evolution, using predicted visual, structured, or latent representations to inform planning \cite{wang2024drivewm,gao2024vista,li2025wote}. Such future evidence helps evaluate the consequences of alternative maneuvers, but prediction fidelity alone does not establish the quality of a candidate plan. The vehicle must also execute the chosen motion within road boundaries under its tracking controller and dynamics. Near a road edge, tracking error and vehicle footprint can turn a seemingly acceptable nominal path into an infeasible execution. Planning benchmarks and closed-loop simulators therefore evaluate executed behavior rather than prediction quality alone \cite{dauner2023pdm,dauner2024navsim,yang2023unisim}. This motivates explicit execution checks alongside predictive scene features.

Existing world-model planners generate candidate trajectories and select a plan through future-aware scoring \cite{li2025wote,gui2026worlddrive}, as illustrated in Figure~\ref{fig:1}(a). A distilled rewarder supplies predictive evidence without generating future scenes online. However, candidate evaluation leaves two execution gaps. First, future-aware selection does not establish road compliance under tracking. Second, re-ranking is bounded by candidate coverage: even a perfect ordering cannot return a feasible trajectory when the retained pool contains none. Our diagnostic on 12,146 NAVSIM v1 scenarios identifies 278 requiring intervention, including 106 (38.1\%) without a road-compliant in-pool alternative. These cases require beyond-pool correction. Physical search can find such corrections, but repeated rollouts are costly. Learning to imitate search introduces a further difficulty: a teacher trajectory that barely passes the execution constraints leaves little tolerance for student prediction errors. Low action error alone may then fail to preserve the teacher's feasibility. This motivates clearance-aware demonstrations and a future-conditioned policy verified with one rollout.

In this paper, we propose Physics-Guided Search Distillation (PGSD), combining execution-aware selection with our search teacher and repair student (Figure~\ref{fig:1}(b,c)). Execution-Aware Selection (EAS) first checks tracked vehicle footprints and selects a road-compliant in-pool trajectory when available. For triggered cases without a feasible alternative, our offline Clearance-Aware Search Teacher (CAST) searches for lateral corrections and prioritizes larger minimum executed road-boundary clearance. The larger target margin buffers imitation errors under the same verifier. Our Future-Conditioned Repair Student (FCRS) then maps frozen rewarder features and physical execution feedback to one beyond-pool correction. The future features supply candidate-conditioned scene context, while execution feedback describes tracking errors and road clearance. PGSD thus transfers inherited predictive knowledge into corrective action while keeping the planner, rewarder, and pretrained physical branch frozen; only the added future-conditioned branch is learned. Single-Rollout Verification (SRV) checks the predicted correction and retains the EAS reference if verification fails. With single-view visual input and map-based execution geometry, PGSD achieves 90.97 PDMS on NAVSIM v1 and 38.96 EPDMS on NAVSIM v2 navhard. The student retains 73.2\% of the search teacher's incremental gain over EAS while replacing up to ten search rollouts with one verification rollout; HUGSIM provides complementary qualitative zero-shot examples.

Our main contributions are summarized as follows:
\begin{itemize}[leftmargin=*,topsep=0pt,partopsep=0pt,itemsep=0pt,parsep=0pt]
\frenchspacing
\item \textbf{Execution-aware selection and beyond-pool repair.} We address both executable candidate selection and the absence of a feasible in-pool plan, mapping inherited future features and physical feedback to a verified correction.
\item \textbf{Clearance-aware search distillation.} We select teacher corrections by minimum executed road-boundary clearance, providing a geometric buffer against imitation errors and distilling search into one student prediction.
\item \textbf{Single-view planning with fixed repair cost.} Two NAVSIM benchmarks demonstrate improved planning with one additional verification rollout, while controlled ablations isolate clearance supervision, execution feedback, and future conditioning.
\end{itemize}

%% file: sections/02_related_work.tex
\section{Related work}\label{sec:2}

\noindent\textbf{End-to-end trajectory planning.}
UniAD \cite{hu2023uniad} and VAD \cite{jiang2023vad} organize perception and prediction around planning, while VADv2 \cite{jiang2026vadv2} and SparseDrive \cite{sun2025sparsedrive} explore probabilistic and sparse representations. To capture multimodality, GenAD \cite{zheng2024genad} models joint future motion in a structured latent space; PLUTO \cite{cheng2024pluto} improves imitation-based planning through architecture, auxiliary objectives, and augmentation. DiffusionDrive \cite{liao2025diffusiondrive}, GoalFlow \cite{xing2025goalflow}, and Diffusion Planner \cite{zheng2025diffusionplanner} use diffusion or flow models to generate multiple candidate trajectories. However, even a diverse candidate pool may contain no executable trajectory. PGSD checks candidate execution and learns beyond-pool corrections to address this coverage limitation.

\noindent\textbf{World models for driving.}
GAIA-1 \cite{hu2023gaia}, DriveDreamer \cite{wang2024drivedreamer}, and DriveDreamer-2 \cite{zhao2025drivedreamer2} learn controllable driving video generation. The predictive GenAD \cite{yang2024genad} and Vista \cite{gao2024vista} improve generalization and controllability, with Vista also supporting action evaluation. Drive-WM \cite{wang2024drivewm} combines visual forecasting with tree-based evaluation, LAW \cite{li2025law} learns latent future representations, and WoTE \cite{li2025wote} scores predicted BEV states. DriveWorld-VLA \cite{jia2026driveworldvla} integrates latent world modeling with vision--language--action planning. World4Drive \cite{zheng2025world4drive} and WorldDrive \cite{gui2026worlddrive} use intention- or trajectory-conditioned latent futures; to reduce repeated prediction cost, WorldDrive distills these into a Future-aware Rewarder (FAR). World4Scorer \cite{pei2026world4scorer} and SafeDrive \cite{kim2026safedrive} refine outcome-based scoring and agent-level risk assessment. However, predictive features used to score candidates do not directly specify a correction under execution constraints. PGSD maps frozen FAR features and execution feedback to verified corrections.

\begin{figure*}[!t]
\centering
\includegraphics[width=0.96\textwidth]{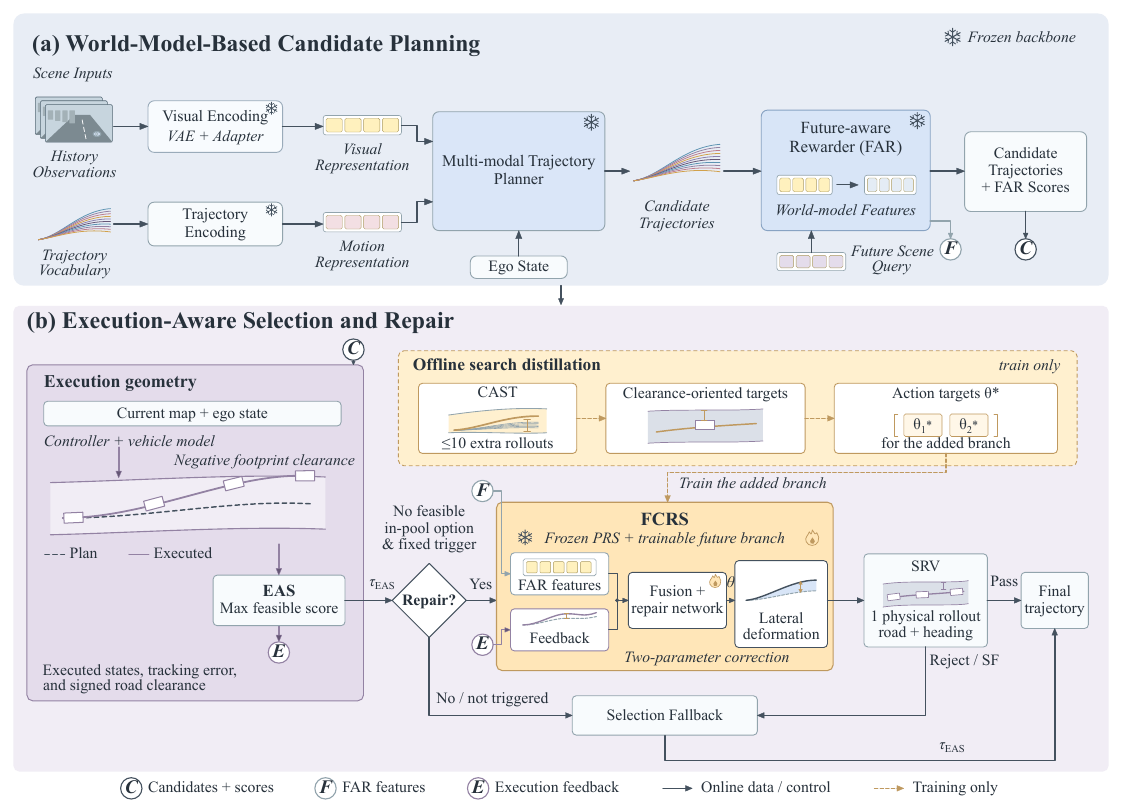}
\caption{\textbf{Overview of PGSD.} $\mathsf C$ denotes candidate trajectories and FAR scores; $\mathsf F$, reused future features; and $\mathsf E$, reference execution feedback. EAS intervenes when executed penetration reaches 5 cm and selects the highest-scoring candidate with nonnegative clearance when available. The selected reference $\tau_{\mathrm{EAS}}$ conditions repair and serves as fallback. FCRS handles triggered beyond-pool repair; SRV verifies its proposal, with Selection Fallback (SF) on rejection. Snowflakes mark frozen networks; flames mark the trainable future branch. Dashed warm-gold arrows denote offline CAST supervision.}
\label{fig:2}
\end{figure*}

\noindent\textbf{Trajectory refinement and distillation.}
Trajectory refinement improves proposals, while distillation yields compact policies. Think Twice Before Driving \cite{jia2023thinktwice} learns offsets using scene features and action-conditioned future representations. ResAD \cite{zheng2026resad}, GuideFlow \cite{liu2026guideflow}, and MeanFuser \cite{wang2026meanfuser} respectively study residual trajectory modeling, constraint-guided generation, and adaptive reconstruction. Diffusion Policy \cite{chi2023diffusionpolicy} models multimodal actions, and consistency models \cite{song2023consistency} distill iterative generation into few-step prediction. Hydra-MDP \cite{li2024hydramdp} and DistillDrive \cite{yu2025distilldrive} transfer planning supervision to compact students. However, these advances do not directly resolve which corrective search targets tolerate prediction errors under a fixed execution verifier. PGSD selects targets by executed clearance and distills corrective search into a future-conditioned repair student. Teacher and student share the same controller, vehicle model, and acceptance rule. Executed-clearance supervision provides a geometric error buffer that action-space imitation losses do not explicitly encode.

%% file: sections/03_method.tex
\section{Method}\label{sec:3}

PGSD connects frozen WorldDrive planning and FAR \cite{gui2026worlddrive} to physical execution through three components (Figure~\ref{fig:2}): execution-aware candidate selection, future-conditioned beyond-pool repair, and clearance-aware search distillation.

\subsection{Execution-aware selection}\label{sec:3.2}\label{sec:3.3}\label{sec:3.3.1}

The planner retains its Top-5 refined trajectories $\tau_i$, indexed by $\mathcal I$. Frozen FAR supplies scores $s_i$ and future features $\mathbf f_i$:
\begin{equation}
\mathsf C=\{(\tau_i,s_i)\}_{i\in\mathcal I},\qquad
\mathsf F=(\mathbf f_i)_{i\in\mathcal I}.
\label{eq:4}
\end{equation}
EAS uses $\mathsf C$; FCRS reuses $\mathsf F$ without online diffusion sampling. Appendices~\ref{app:backbone} and~\ref{app:far} detail planner pretraining.

For ego state $\mathbf e_0$ and controller--vehicle configuration $\boldsymbol\xi$, we evaluate ego footprints against drivable area $\mathcal D$:
\begin{equation}
\begin{aligned}
\widehat{\mathbf X}_i&=\mathcal R(\tau_i,\mathbf e_0;\boldsymbol\xi),\\
d_{i,k}&=\operatorname{SignedClearance}\!\left(\operatorname{Footprint}(\widehat{\mathbf x}_{i,k}),\mathcal D\right),\\
m_i&=\min_k d_{i,k},\\
\mathsf E_i&=\left(\widehat{\mathbf X}_i,\operatorname{TrackError}(\widehat{\mathbf X}_i,\tau_i),\{d_{i,k}\}_k\right).
\end{aligned}
\label{eq:6}
\end{equation}
Here, $k=0,\ldots,N_e$ includes the initial pose. Clearance is the minimum over four ego-footprint corners, positive inside and negative outside the road. With $b=\arg\max_{i\in\mathcal I}s_i$ and $h=\mathbb I[m_b\leq-0.05]$, EAS intervenes when the executed footprint crosses the road boundary by 5 cm:
\begin{equation}
\begin{aligned}
\mathcal I^{+}&=\{i\in\mathcal I\mid m_i\geq0\},\quad
r=h\,\eta_{\mathrm{valid}}\,\mathbb I[\mathcal I^{+}=\varnothing],\\
i_{\mathrm{EAS}}&=\begin{cases}
\arg\max_{i\in\mathcal I^{+}}s_i,&h=1,\ \mathcal I^{+}\neq\varnothing,\\
b,&\text{otherwise}.
\end{cases}
\end{aligned}
\label{eq:7}
\end{equation}
The validity flag $\eta_{\mathrm{valid}}$ requires an available map and finite execution states; ties follow candidate order. On a threshold violation, EAS selects the highest-scoring feasible alternative; an empty feasible set triggers repair only with valid feedback. EAS changes only the candidate index, preserving each proposal's original geometry. The reference $\tau_{\mathrm{EAS}}=\tau_{i_{\mathrm{EAS}}}$ supplies execution feedback and serves as fallback if the repair fails verification. Appendix~\ref{app:execution} specifies the controller, vehicle model, and rollout settings.

\subsection{Future-conditioned repair student}\label{sec:3.3.2}\label{sec:3.3.4}

On a repair trigger ($r=1$), FCRS combines the physical descriptor $\mathbf u_E=(\mathbf X_E,\mathbf v_E)$ with the EAS reference's FAR feature. A pretrained Physics-Only Repair Student (PRS) encodes nominal and executed motion, tracking errors, and road clearance into a 64-dimensional representation $\mathbf z_g$ and a physical correction. FCRS adds a learned residual:
\begin{equation}
\begin{aligned}
\boldsymbol\theta_{\mathrm{phys}}&=G_{\mathrm{phys}}(\mathbf u_E;\boldsymbol\omega_0),\\
\mathbf z_F&=A_{\boldsymbol\phi}(\mathbf f_{i_{\mathrm{EAS}}}),\\
\widehat{\boldsymbol\theta}_{\boldsymbol\phi}&=\boldsymbol\theta_{\mathrm{phys}}+H_{\boldsymbol\phi}([\mathbf z_g;\mathbf z_F]).
\end{aligned}
\label{eq:10}
\end{equation}
Only the future-projection and residual-MLP parameters $\boldsymbol\phi$ are trained; the planner, FAR, and PRS remain frozen. FAR conditions both training and inference. Appendices~\ref{app:descriptor} and~\ref{app:fcrs} specify the descriptor and architecture.

For waypoint time $t\in\{0.5,1.0,\ldots,4.0\}$, let $u_t=t/4$ and $\mathbf n_t$ be the reference heading's unit normal. Two parameters control terminal and intermediate lateral displacement:
\begin{equation}
\begin{aligned}
\boldsymbol\theta&=\Pi_{\Theta}(\widehat{\boldsymbol\theta}_{\boldsymbol\phi}),\\
\delta_t&=\theta_1u_t^2+\tfrac{27}{4}\theta_2u_t^2(1-u_t),\\
\mathbf p_{\mathrm{rep},t}&=\mathbf p_{\mathrm{EAS},t}+\delta_t\mathbf n_t,\\
\tau_{\mathrm{rep}}&=\mathcal D_{\mathrm{lat}}(\tau_{\mathrm{EAS}},\boldsymbol\theta).
\end{aligned}
\label{eq:11}
\end{equation}
Projection uniformly rescales the parameters to enforce $\max_t|\delta_t|\leq0.75$ m. Heading is updated by the change in path tangent relative to the reference; points with reference tangent speed below 0.5 m/s retain their original heading. Zero deformation preserves the reference.

\noindent\textbf{Single-rollout verification.} SRV rejects nonfinite proposals or execution states. Acceptance requires at least 2 cm road clearance at every rollout state, including the initial pose, and an absolute wrapped heading difference of at most 0.20 rad from the EAS reference at each waypoint. Let $a$ denote these checks and the bound in Eq.~\eqref{eq:11}:
\begin{equation}
\tau_{\mathrm{out}}=\begin{cases}
\tau_{\mathrm{rep}},&r=1,\ a(\tau_{\mathrm{rep}})=1,\\
\tau_{\mathrm{EAS}},&\text{otherwise}.
\end{cases}
\label{eq:16}
\end{equation}
Each triggered attempt uses one proposal and one additional rollout; rejection or no trigger retains the EAS reference.

\subsection{Clearance-aware search distillation}\label{sec:3.3.3}

CAST searches $\tau(\boldsymbol\theta)=\mathcal D_{\mathrm{lat}}(\tau_{\mathrm{EAS}},\boldsymbol\theta)$ from zero deformation, balancing clearance shortfall, correction magnitude, and smoothness. It reuses the reference rollout and queries $\Theta_B\subseteq\Theta$ with $|\Theta_B|\leq B=10$; Appendix~\ref{app:execution} gives the objective and finite differences. Qualifying proposals satisfy the same verifier as the student:
\begin{equation}
\Theta_B^{+}=\{\boldsymbol\theta\in\Theta_B\mid a(\tau(\boldsymbol\theta))=1\}.
\label{eq:12}
\end{equation}
Let $\mathcal B_k(\boldsymbol\theta)$ be the executed ego footprint of $\tau(\boldsymbol\theta)$ at state $k$. For $\Theta_B^{+}\neq\varnothing$, CAST maximizes executed clearance:
\begin{equation}
\begin{aligned}
c(\boldsymbol\theta)&=\min_{0\leq k\leq N_e}\operatorname{SignedClearance}\!\left(\mathcal B_k(\boldsymbol\theta),\mathcal D\right),\\
\boldsymbol\theta^{*}&\in\arg\max_{\boldsymbol\theta\in\Theta_B^{+}}c(\boldsymbol\theta).
\end{aligned}
\label{eq:13}
\end{equation}
Solutions within $10^{-4}$ m of the maximum are ranked by smaller RMS lateral deformation and then stable proposal order. Clearance reuses the search rollouts; larger target margins buffer imitation error.

PRS is pretrained on these demonstrations using the same losses as FCRS. FCRS then freezes PRS, the planner, and FAR and learns $\boldsymbol\phi$ using teacher-repaired examples $\mathcal D_{\mathrm{rep}}$ and already-feasible references $\mathcal D_{\mathrm{keep}}$. Write $\boldsymbol\theta_{\boldsymbol\phi}=\Pi_{\Theta}(\widehat{\boldsymbol\theta}_{\boldsymbol\phi})$, $\tau_{\boldsymbol\phi}=\tau(\boldsymbol\theta_{\boldsymbol\phi})$, and $\tau^{*}=\tau(\boldsymbol\theta^{*})$. The objective is
\begin{equation}
\begin{aligned}
\mathcal L_{\mathrm{rep}}(\boldsymbol\phi)&=\mathcal L_{\mathrm{act}}+0.25\mathcal L_{\mathrm{plan}}+0.10\mathcal L_{\mathrm{keep}},\\
\mathcal L_{\mathrm{act}}&=\mathbb E_{\mathcal D_{\mathrm{rep}}}\!\left[\ell_H\!\left(\tfrac{\boldsymbol\theta_{\boldsymbol\phi}-\boldsymbol\theta^{*}}{s_{\theta}}\right)\right],\\
\mathcal L_{\mathrm{plan}}&=\mathbb E_{\mathcal D_{\mathrm{rep}}}\!\left[\ell_{\tau}(\tau_{\boldsymbol\phi},\tau^{*})\right],\\
\mathcal L_{\mathrm{keep}}&=\mathbb E_{\mathcal D_{\mathrm{keep}}}\!\left[\left\|\boldsymbol\theta_{\boldsymbol\phi}/s_{\theta}\right\|_2^2\right].
\end{aligned}
\label{eq:14}
\end{equation}
Here, $\ell_H$ is mean Huber loss with unit transition and $s_{\theta}=0.5$. The trajectory loss sums mean squared position errors normalized by 0.75 m and mean squared wrapped heading errors normalized by 0.20 rad. Preservation penalizes deformation of feasible references.

%% file: sections/04_experiments.tex
\section{Experiments}\label{sec:4}

\subsection{Experimental setup}\label{sec:4.1}

\noindent\textbf{Datasets.} We evaluate on all 12,146 NAVSIM v1 navtest scenarios \cite{dauner2024navsim} and NAVSIM v2 navhard\_two\_stage (5,912 frames, 225 groups) \cite{cao2025pseudosimulation}. CAST demonstrations on navtrain are split by recording session, keeping references and their proposals together. Repair ablations use 241 execution-infeasible development trajectories, including a teacher-feasible subset of 202; full-planning scores cover the complete test splits. HUGSIM Hard \cite{zhou2026hugsim} provides complementary qualitative zero-shot examples.

\input{tables/table1}

\noindent\textbf{Evaluation metrics.} We report PDMS for four-second non-reactive NAVSIM v1 rollouts and aggregate EPDMS under NAVSIM v2's official reactive two-stage protocol. Scores use the 0--100 scale; Appendix~\ref{app:metrics} defines their components and aggregation. Repair evaluation additionally reports proposal qualification, minimum executed clearance, and rejection rates under the same verifier (Section~\ref{sec:3.3.4}); fallback is not successful repair. Native planning metrics assess collision and TTC outcomes. Repair cost counts additional rollouts per trigger and cached single-thread CPU latency, excluding shared planning and EAS.

\noindent\textbf{Implementation details.} We train in PyTorch on one NVIDIA GeForce RTX 5090, freezing the pretrained WorldDrive planner/FAR and PRS while learning the future projection and residual predictor. The visual backbone uses a single camera view; execution geometry additionally uses benchmark ego states and drivable-area polygons with a tracking controller and vehicle model. FCRS reuses FAR features without additional rewarder computation and verifies one proposal per trigger; CAST uses up to ten additional rollouts. Appendix~\ref{app:implementation} details training and execution settings.

\input{tables/table2}

\begin{figure}[!t]
\centering
\includegraphics[width=0.96\columnwidth]{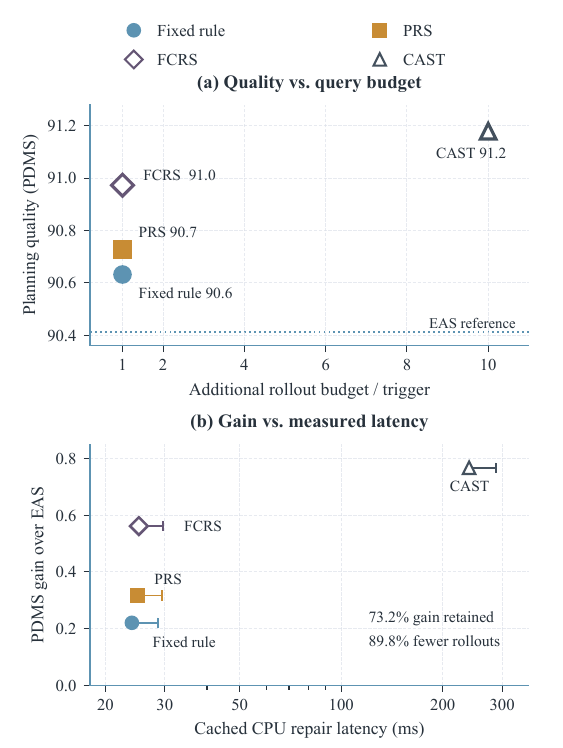}
\caption{\textbf{Planning quality and additional repair cost.} (a) PDMS versus additional rollouts per trigger; the horizontal line marks EAS. (b) Gain over EAS versus cached repair latency.}
\label{fig:3}
\end{figure}

\subsection{Comparison with state-of-the-art methods}\label{sec:4.2}

To compare planning performance under common benchmark protocols, we evaluate on NAVSIM v1 navtest and NAVSIM v2 navhard using each benchmark's native metric. Table~\ref{tab:1} groups v1 methods by sensor configuration, retaining published baseline scores; PGSD additionally uses map geometry for execution checks. Our \textbf{single-view} student reaches \textbf{91.0 PDMS}, exceeding the strongest listed prior single-view planner, WorldDrive, by \textbf{2.0 points} and the listed multi-view/LiDAR WoTE by 2.7 points. Against WorldDrive, DAC rises from 96.8 to 98.9 and EP from 83.3 to 84.9, while TTC and comfort remain at 95.2 and 100.0. The search teacher is reported separately as a reference; student inference verifies one repair proposal per trigger without iterative online search.

On NAVSIM v2, Table~\ref{tab:2} follows the same reactive two-stage protocol for all listed methods; the WorldDrive baseline is reproduced using candidate-score selection. With \textbf{single-view} visual input and map-based execution checks, PGSD reaches \textbf{38.96 EPDMS}, surpassing WorldDrive by \textbf{11.24 points (40.5\%)} and the reported DiffusionDrive result by 11.46 points. Relative to WorldDrive, DAC increases by 13.1 points in Stage 1 and 13.0 in Stage 2, alongside EP gains of 0.8 points in both stages. Joint DAC and EP gains across both protocols indicate improved road compliance while preserving forward progress.

\begin{figure*}[!t]
\centering
\includegraphics[width=\textwidth,trim=0 10 0 4,clip]{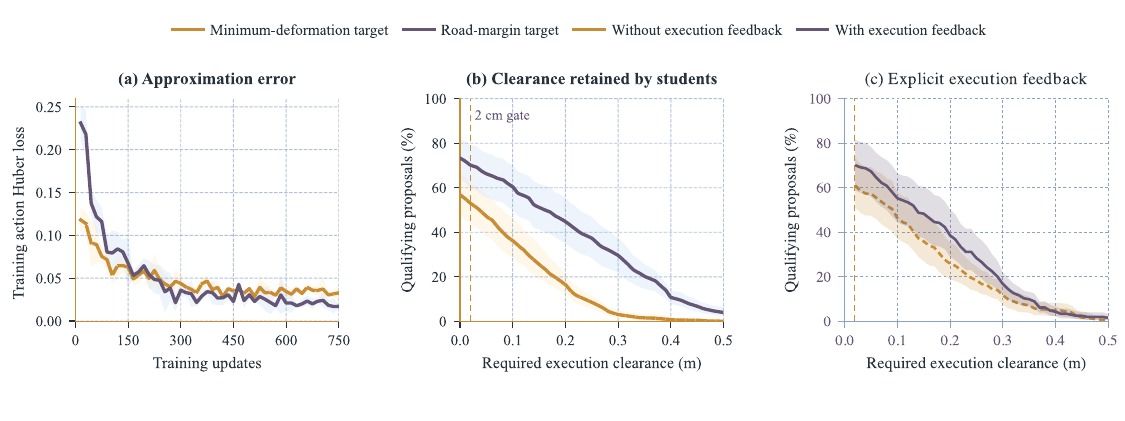}
\caption{\textbf{Demonstration clearance and execution feedback.} (a,b) Demonstration selection with full execution feedback (three paired seeds): (a) training action Huber loss (mean $\pm$ sample SD); (b) qualifying rates versus required clearance, averaged over seeds. (c) Effect of execution feedback under road-margin supervision. Panels (b,c) evaluate the same 241 development trajectories.}
\label{fig:4}
\end{figure*}

\begin{figure*}[!t]
\centering
\includegraphics[width=\textwidth]{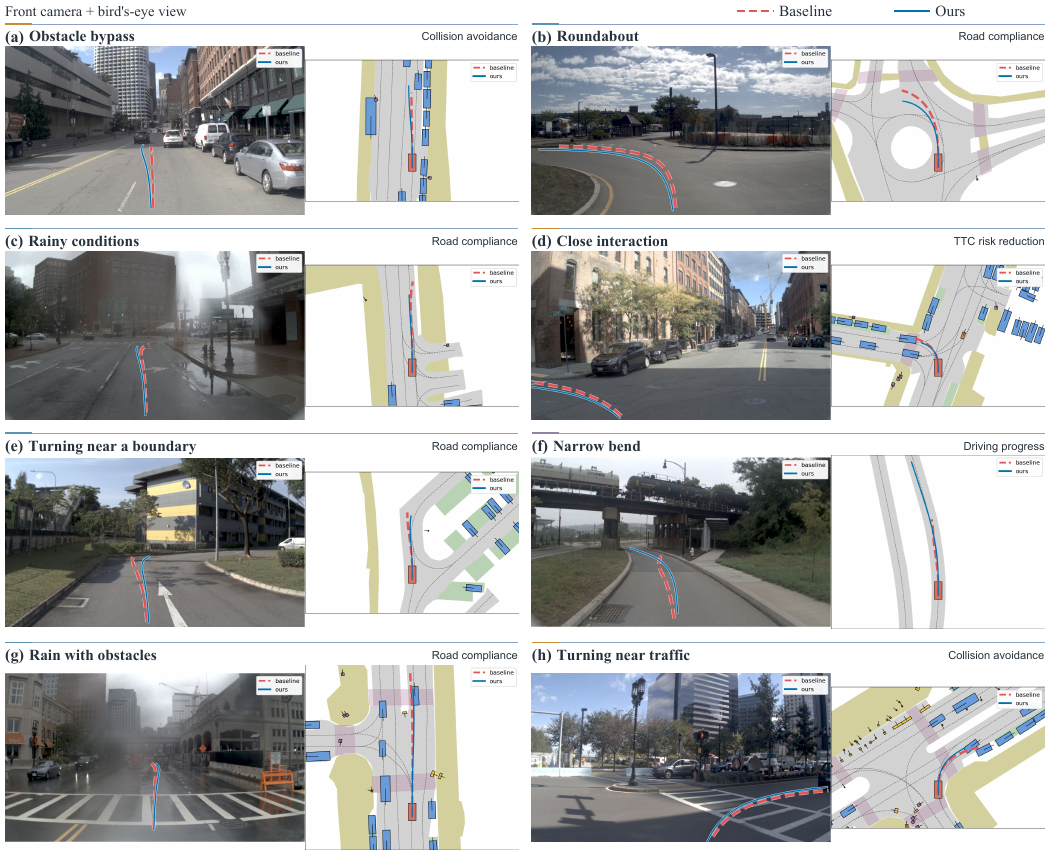}
\caption{\textbf{Execution-aware selection on NAVSIM.} Each panel pairs a front-camera view with a bird's-eye view of four-second simulated execution. Red dashed lines denote WorldDrive; blue solid lines denote execution-aware selection. Panels show (a) obstacle bypassing, (b) a roundabout, (c) rainy conditions, (d) close interaction, (e) turning near a boundary, (f) a narrow bend, (g) rain with obstacles, and (h) turning near traffic. Surrounding actors are shown at the observation time.}
\label{fig:5}
\end{figure*}

\subsection{Ablation study}\label{sec:4.3}

\noindent\textbf{Effect of selection and repair components.}
We first isolate selection and repair under a common WorldDrive backbone (Table~\ref{tab:4}). EAS raises PDMS from 89.0 to 90.4 by choosing a road-compliant alternative within the retained pool, establishing the contribution of execution-aware selection alone. Fixed-rule correction and PRS further reach 90.6 and 90.7, respectively, showing that beyond-pool correction adds gains after re-ranking. PRS raises DAC from 98.4 to 98.8 and EP from 84.6 to 84.9 relative to EAS. These changes indicate that repair improves boundary compliance while preserving progress. Since EAS cannot generate a new trajectory, the additional gains show the value of repair when the retained pool contains no feasible trajectory.

\input{tables/table4}

\noindent\textbf{Effect of future conditioning.}
FCRS augments frozen PRS with a trainable FAR-feature projection and residual predictor, increasing PDMS from 90.7 to 91.0. Both variants retain FAR for candidate scoring, use identical execution constraints, and verify one correction per trigger. The comparison therefore measures the added future-conditioned repair branch rather than the presence of a world-model scorer. Table~\ref{tab:3} shows a gain of 0.56 PDMS over EAS for FCRS versus 0.32 for PRS, computed from unrounded scores. Future features complement physical feedback: the latter describes how the reference tracks and approaches road boundaries, while the former supplies candidate-conditioned scene context for selecting a correction. This supports combining scene context and physical feedback at fixed rollout cost.

\noindent\textbf{Effect of clearance-aware demonstrations.}
We compare physics-only students with the same search proposals, architecture, training budget, full execution feedback, and three paired seeds, varying only target selection. Minimum-deformation targets choose the qualifying correction with the smallest RMS lateral displacement; road-margin targets prioritize executed clearance. Both cover the same 680 of 773 infeasible training references, while median target clearance increases from 9.8 to 31.3 cm. On all 241 development references, repair success increases from \textbf{$53.0\pm1.0\%$ to $70.1\pm1.1\%$}, a gain of 17.15 percentage points; rejection falls from 47.0\% to 29.9\%. On the same 202 teacher-feasible references, success rises from 62.7\% to 82.7\%. Because target coverage is unchanged, the gain supports the quality of clearance-aware supervision rather than access to additional repairable examples.

\noindent\textbf{Why clearance matters for imitation.}
Both target policies converge in action Huber loss (Figure~\ref{fig:4}(a)), but similar convergence does not establish equivalent execution feasibility. The road-margin student qualifies more often across clearance thresholds (Figure~\ref{fig:4}(b)), consistent with larger target margins absorbing prediction errors. Reference-averaged median executed clearance rises from 4.32 to 16.45 cm, with improvements on 228 of 241 references (94.6\%; Appendix~\ref{app:clearance}). Recorded clearance rejections fall from 331/723 to 206/723 proposals, whereas heading rejections change from 9/723 to 10/723 (Appendix~\ref{app:rejections}). This concentrates the benefit in the constraint targeted by the teacher. Negative-clearance cases remain, motivating verification and fallback even with improved supervision.

\noindent\textbf{Effect of explicit execution feedback.}
Under the same road-margin supervision, we hold the student architecture, training budget, and verification procedure fixed, removing executed states, tracking errors, and executed-clearance inputs together. Repair success falls from \textbf{70.1\% to 62.38\%} on the same 241 references (Figure~\ref{fig:4}(c)), a decrease of 7.72 percentage points. Nominal motion and road geometry thus leave useful information unobserved: controller response can move the tracked footprint toward a boundary even when the nominal path appears compliant. The controlled removal supports execution feedback as a group; it does not isolate individual channels. Together with the demonstration study, this shows that both the corrective target and the physical inputs affect repair reliability.

\noindent\textbf{Distillation quality and repair cost.}
We quantify PDMS gain over EAS and retention relative to CAST:
\begin{equation}
\begin{aligned}
\Delta_m&=\mathrm{PDMS}_m-\mathrm{PDMS}_{\mathrm{EAS}},\\
R_m&=\Delta_m/\Delta_{\mathrm{CAST}}.
\end{aligned}
\label{eq:19}
\end{equation}
FCRS gains 0.56 points versus CAST's 0.77, retaining \textbf{73.2\%} of the teacher's gain from unrounded scores (Table~\ref{tab:3}). Fixed-rule correction and PRS retain 28.7\% and 41.3\%. FCRS reaches 90.97 PDMS, within 0.21 points of CAST, showing that the stronger learned correction narrows the quality gap without increasing rollout queries. In a paired comparison of 106 repair triggers, search uses 1,040 additional rollouts versus 106 for PRS, an 89.8\% reduction; FCRS follows the same one-verification design. Median cached repair latency is 25.13 ms for FCRS, 24.85 ms for PRS, and 239.49 ms for search. Figure~\ref{fig:3} illustrates this quality--cost trade-off; these timings measure the correction stage with cached inputs, rather than total planner runtime.

\input{tables/table3}

\subsection{Qualitative analysis}\label{sec:4.4}

Figure~\ref{fig:5} examines the role of EAS through eight NAVSIM scenarios. Front-camera views provide scene context, and bird's-eye views show four-second simulated execution. At the roundabout, turning boundary, and rainy scenes, EAS selects trajectories that remain within the drivable corridor where WorldDrive crosses road boundaries. The bypassing and turning-near-traffic cases also improve collision outcomes, while the close-interaction example improves road compliance and TTC. The narrow bend illustrates compliant forward progress in constrained geometry. These cases explain how execution-aware selection exploits alternatives already in the candidate pool, alongside the beyond-pool repair evaluated in Section~\ref{sec:4.3}.

\textbf{Qualitative zero-shot planning on HUGSIM.} We evaluate PGSD on HUGSIM Hard \cite{zhou2026hugsim} without simulator-specific training or fine-tuning. Plans follow narrow and curved corridors and adapt to nearby vehicles. More visualizations are provided in Appendix~\ref{app:hugsim} (Figure~\ref{fig:s5}).

%% file: tables/table1.tex
\begin{table*}[!t]
\centering
\caption{\textbf{Planning performance on NAVSIM v1 navtest.} Baseline scores follow WorldDrive \cite{gui2026worlddrive}. SV/MV and L denote single-/multi-view cameras and LiDAR. PGSD also uses road maps. $\Delta$PDMS is the absolute gain over the reported WorldDrive (89.0). Bold/underline mark best/second-best distinct scores, including ties; CAST is a search reference.}
\label{tab:1}
\begingroup
\small
\setlength{\tabcolsep}{2.5pt}
\renewcommand{\arraystretch}{1.12}
\begin{tabular}{@{}>{\raggedright\arraybackslash}p{\dimexpr\textwidth-400pt\relax}>{\centering\arraybackslash}p{43pt}>{\centering\arraybackslash}p{58pt}*{3}{>{\raggedleft\arraybackslash}p{34pt}}>{\raggedleft\arraybackslash}p{40pt}>{\raggedleft\arraybackslash}p{32pt}>{\raggedleft\arraybackslash}p{40pt}>{\raggedleft\arraybackslash}p{40pt}@{}}
\toprule[1pt]
Method & Sensors & \mbox{World model} & NC $\uparrow$ & DAC $\uparrow$ & TTC $\uparrow$ & Comf. $\uparrow$ & EP $\uparrow$ & PDMS $\uparrow$ & $\Delta$PDMS \\
\midrule
TransFuser~\cite{chitta2023transfuser} & MV \& L & $\times$ & 97.7 & 92.8 & 92.8 & \textbf{100.0} & 79.2 & 84.0 & -5.0  \\
Hydra-MDP~\cite{li2024hydramdp} & MV \& L & $\times$ & 98.3 & 96.0 & 94.6 & \textbf{100.0} & 78.7 & 86.5 & -2.5  \\
DiffusionDrive~\cite{liao2025diffusiondrive} & MV \& L & $\times$ & 98.2 & 96.2 & 94.7 & \textbf{100.0} & 82.2 & 88.1 & -0.9  \\
WoTE~\cite{li2025wote} & MV \& L & $\checkmark$ & 98.5 & 96.8 & 94.9 & \underline{99.9} & 81.9 & 88.3 & -0.7  \\
\midrule[0.3pt]
UniAD~\cite{hu2023uniad} & MV & $\times$ & 97.8 & 91.9 & 92.9 & \textbf{100.0} & 78.8 & 83.4 & -5.6  \\
PARA-Drive~\cite{weng2024paradrive} & MV & $\times$ & 97.9 & 92.4 & 93.0 & 99.8 & 79.3 & 84.0 & -5.0  \\
LAW~\cite{li2025law} & MV & $\checkmark$ & 96.4 & 95.4 & 88.7 & \underline{99.9} & 81.7 & 84.6 & -4.4  \\
World4Drive~\cite{zheng2025world4drive} & MV & $\checkmark$ & 97.4 & 94.3 & 92.8 & \textbf{100.0} & 79.9 & 85.1 & -3.9  \\
FSDrive~\cite{zeng2025futuresightdrive} & MV & $\checkmark$ & 98.2 & 93.8 & 93.3 & \underline{99.9} & 80.1 & 85.1 & -3.9  \\
\midrule[0.3pt]
DrivingGPT~\cite{chen2024drivinggpt} & SV & $\checkmark$ & \textbf{98.9} & 90.7 & 94.9 & 95.6 & 79.7 & 82.4 & -6.6  \\
Epona~\cite{zhang2025epona} & SV & $\checkmark$ & 97.9 & 95.1 & 93.8 & \underline{99.9} & 80.4 & 86.2 & -2.8  \\
ImagiDrive~\cite{li2026imagidrive} & SV & $\checkmark$ & 98.6 & 96.2 & 94.5 & \textbf{100.0} & 80.5 & 87.4 & -1.6  \\
PWM~\cite{zhao2025pwm} & SV & $\checkmark$ & 98.6 & 95.9 & \underline{95.4} & \textbf{100.0} & 81.8 & 88.1 & -0.9  \\
DriveVLA-W0~\cite{li2025drivevlaw0} & SV & $\checkmark$ & \underline{98.7} & 96.2 & \textbf{95.5} & \textbf{100.0} & 82.2 & 88.4 & -0.6  \\
WorldDrive~\cite{gui2026worlddrive} & SV & $\checkmark$ & 98.4 & 96.8 & 95.2 & \textbf{100.0} & 83.3 & 89.0 & +0.0  \\
\midrule[0.8pt]
\textbf{CAST (Ours)} & \textbf{SV} & $\checkmark$ & \textbf{98.9} & \textbf{99.0} & 95.3 & \textbf{100.0} & \textbf{85.0} & \textbf{91.2} & +2.2  \\
\rowcolor{studentrow}
\textbf{FCRS (Ours)} & \textbf{SV} & $\checkmark$ & \underline{98.7} & \underline{98.9} & 95.2 & \textbf{100.0} & \underline{84.9} & \underline{91.0} & \textbf{+2.0}  \\
\bottomrule[1pt]
\end{tabular}
\endgroup
\end{table*}

%% file: tables/table2.tex
\begin{table}[!t]
\centering
\caption{\textbf{Planning performance on NAVSIM v2 navhard.} S1 and S2 denote the two evaluation stages; EPDMS is the aggregate two-stage score. SV/MV denote single-/multi-view cameras. $\dagger$ Results reported in WorldDrive \cite{gui2026worlddrive}. We reproduce the WorldDrive baseline using its candidate-score selection. All scores use the 0--100 scale. Bold/underline indicate the best/second-best distinct scores per stage and aggregate, including ties.}
\label{tab:2}
\begingroup
\scriptsize
\setlength{\tabcolsep}{1.8pt}
\renewcommand{\arraystretch}{1.12}
\resizebox{\columnwidth}{!}{%
\begin{tabular}{@{}lccrrrrrc@{}}
\toprule[1pt]
Method & \makebox[13pt][r]{Sensors} & Stage & NC $\uparrow$ & DAC $\uparrow$ & EP $\uparrow$ & TTC $\uparrow$ & HC $\uparrow$ & EPDMS $\uparrow$ \\
\midrule
\multirow{2}{*}{LTF$\dagger$} & \multirow{2}{*}{MV} & S1 & \underline{96.2} & 79.5 & \textbf{84.1} & 95.1 & \underline{97.5} & \multirow{2}{*}{23.1} \\
 &  & S2 & 77.7 & 70.2 & \underline{85.1} & 75.6 & 95.7 &  \\
\midrule[0.3pt]
\multirow{2}{*}{DiffusionDrive$\dagger$} & \multirow{2}{*}{MV} & S1 & \textbf{96.8} & \underline{86.0} & \underline{84.0} & \textbf{95.8} & \textbf{97.6} & \multirow{2}{*}{27.5} \\
 &  & S2 & 80.1 & \underline{72.8} & \textbf{85.9} & 76.6 & 96.3 &  \\
\midrule[0.3pt]
\multirow{2}{*}{WorldDrive} & \multirow{2}{*}{SV} & S1 & 95.7 & 80.4 & 81.8 & 94.4 & \textbf{97.6} & \multirow{2}{*}{\underline{27.72}} \\
 &  & S2 & \underline{82.9} & 68.0 & 81.6 & \underline{78.2} & \underline{97.4} &  \\
\midrule[0.8pt]
\rowcolor{studentrow}
 &  & S1 & \underline{96.2} & \textbf{93.5} & 82.6 & \underline{95.5} & \textbf{97.6} &  \\
\rowcolor{studentrow}
\multirow{-2}{*}{\textbf{Ours}} & \multirow{-2}{*}{\textbf{SV}} & S2 & \textbf{84.3} & \textbf{81.0} & 82.4 & \textbf{79.7} & \textbf{97.5} & \multirow{-2}{*}{\textbf{38.96}} \\
\bottomrule[1pt]
\end{tabular}
}
\endgroup
\end{table}

%% file: tables/table4.tex
\begin{table}[!htbp]
\centering
\caption{\textbf{Component analysis on NAVSIM v1.} WorldDrive is reproduced. All variants retain FAR for candidate scoring; FAR $\rightarrow$ student denotes FAR features used for repair.}
\label{tab:4}
\begingroup
\footnotesize
\setlength{\tabcolsep}{2pt}
\renewcommand{\arraystretch}{1.12}
\resizebox{\columnwidth}{!}{%
\begin{tabular}{@{}lccccrrr@{}}
\toprule[1pt]
Method & EAS & Repair & \begin{tabular}[c]{@{}c@{}}FAR $\rightarrow$\\student\end{tabular} & SRV & DAC $\uparrow$ & EP $\uparrow$ & PDMS $\uparrow$ \\
\midrule
\multicolumn{8}{@{}l}{\textit{Component variants}} \\[1pt]
WorldDrive & \textemdash{} & None & \textemdash{} & \textemdash{} & 97.0 & 83.5 & 89.0 \\
+ EAS & $\checkmark$ & None & \textemdash{} & \textemdash{} & 98.4 & 84.6 & 90.4 \\
+ Fixed rule & $\checkmark$ & Rule & \textemdash{} & $\checkmark$ & 98.7 & 84.8 & 90.6 \\
+ PRS & $\checkmark$ & Student & \textemdash{} & $\checkmark$ & 98.8 & 84.9 & 90.7 \\
\midrule[0.5pt]
\multicolumn{8}{@{}l}{\textit{Full framework and teacher reference}} \\[1pt]
\rowcolor{studentrow}
+ FCRS & $\checkmark$ & Student & $\checkmark$ & $\checkmark$ & 98.9 & 84.9 & 91.0 \\
CAST & $\checkmark$ & Search & \textemdash{} & $\checkmark$ & \textbf{99.0} & \textbf{85.0} & \textbf{91.2} \\
\bottomrule[1pt]
\end{tabular}
}
\endgroup
\end{table}

%% file: tables/table3.tex
\begin{table}[!htbp]
\centering
\caption{\textbf{Distillation quality and additional repair cost.} Gains and retention ratios use the common EAS reference and CAST, computed before rounding. Rollouts are mean additional queries per repair trigger; FCRS uses a fixed budget of one. Latencies measure the cached single-thread CPU correction stage in milliseconds. Dashes denote inapplicable entries.}
\label{tab:3}
\begingroup
\footnotesize
\setlength{\tabcolsep}{2.6pt}
\renewcommand{\arraystretch}{1.12}
\resizebox{\columnwidth}{!}{%
\begin{tabular}{@{}lrrrrr@{}}
\toprule[1pt]
Method & PDMS $\uparrow$ & \begin{tabular}[c]{@{}c@{}}$\Delta$ vs.\\EAS $\uparrow$\end{tabular} & \begin{tabular}[c]{@{}c@{}}Gain\\retained\\(\%) $\uparrow$\end{tabular} & \begin{tabular}[c]{@{}c@{}}Rollouts\\/ trigger $\downarrow$\end{tabular} & \begin{tabular}[c]{@{}c@{}}Median latency\\(ms) $\downarrow$\end{tabular} \\
\midrule
\multicolumn{6}{@{}l}{\textit{Selection and repair references}} \\[1pt]
EAS & 90.4 & +0.00 & \textemdash{} & 0.00 & \textemdash{} \\
Fixed rule & 90.6 & +0.22 & 28.7 & 1.00 & 23.96 \\
PRS & 90.7 & +0.32 & 41.3 & 1.00 & 24.85 \\
\midrule[0.5pt]
\multicolumn{6}{@{}l}{\textit{Future-conditioned repair and search}} \\[1pt]
\rowcolor{studentrow}
\textbf{FCRS} & \textbf{91.0} & \textbf{+0.56} & \textbf{73.2} & \textbf{1.00} & \textbf{25.13} \\
CAST & 91.2 & +0.77 & 100.0 & 9.81 & 239.49 \\
\bottomrule[1pt]
\end{tabular}
}
\endgroup
\end{table}

%% file: sections/05_conclusion.tex
\section{Conclusion}\label{sec:5}

PGSD addresses two limitations of future-aware planning: accurate prediction does not ensure road-compliant execution, and re-ranking cannot resolve a candidate pool with no feasible trajectory. PGSD combines execution-aware selection with a student that maps frozen FAR features and physical feedback to beyond-pool corrections. Clearance-oriented teacher targets buffer imitation errors, while each repair uses one verification rollout with fallback on rejection. With single-view visual input and map-based execution checks, PGSD achieves 90.97 PDMS on NAVSIM v1, within 0.21 points of CAST while retaining 73.2\% of its incremental gain over EAS. On NAVSIM v2 navhard, it reaches 38.96 EPDMS, improving on WorldDrive by 40.5\%. Ablations support target clearance, execution feedback, and future conditioning; HUGSIM provides complementary qualitative evidence. These results show how inherited predictive features can support efficient corrective planning. Future work will address richer maneuvers and interaction-aware constraints.

%% file: sections/06_llm_usage.tex
\section*{LLM Usage Statement}\label{sec:llm}
OpenAI ChatGPT and Codex were used for language editing, reference checking, manuscript and figure formatting, and boilerplate code completion. The authors reviewed all revisions and take full responsibility for the final content.

%% file: appendix/appendix.tex
\clearpage
\raggedbottom
\setlength{\parskip}{0pt}
\appendix
\setcounter{figure}{0}
\setcounter{table}{0}
\setcounter{equation}{0}
\renewcommand{\thefigure}{S\arabic{figure}}
\renewcommand{\thetable}{S\arabic{table}}
\renewcommand{\theequation}{S\arabic{equation}}
\renewcommand{\thepage}{S\arabic{page}}
\setcounter{page}{1}
\maketitlesupplementary
\input{appendix/00_implementation}
\input{appendix/02_far_origin}
\input{appendix/03a_clearance_distribution_figure}
\input{appendix/01_evaluation}
\input{appendix/03_clearance_statistics}
\raggedcolsend
\clearpage
\onecolumn
\input{appendix/03b_paired_clearance_figure}
\input{appendix/04_rejections}
\clearpage
\input{appendix/05_hugsim}

%% file: appendix/00_implementation.tex
\section{Implementation details}\label{app:implementation}

\subsection{Pretrained candidate planner}\label{app:backbone}

PGSD uses the pretrained WorldDrive stage-2 planner and FAR \cite{gui2026worlddrive}. The causal VAE uses CogVideoX-2B weights. The inherited backbone encodes historical observations $H$, current ego state $\mathbf e_0$, and trajectory vocabulary $\mathcal A$ as follows:
\begin{equation}
\begin{aligned}
\mathbf z^{\mathrm{vis}}&=A_{\mathrm{vis}}\!\left(E_{\mathrm{VAE}}(H)\right),\\
\mathbf Z^{\mathrm{mot}}&=E_{\mathrm{traj}}(\mathcal A),\\
\{(q_i,\Delta\tau_i)\}_{i=1}^{256}&=P\!\left(\mathbf z^{\mathrm{vis}},\mathbf Z^{\mathrm{mot}},\mathbf e_0\right).
\end{aligned}
\label{eq:s-backbone}
\end{equation}
Each anchor contains eight future rear-axle poses over four seconds, with a two-dimensional position and heading per pose. The planner predicts an offset and initial ranking score for each anchor. It refines $\tau_i=\mathbf a_i+\Delta\tau_i$ and retains indices $\mathcal I=\operatorname{TopK}(\{q_i\}_{i=1}^{256})$ with $K=5$. FAR evaluates this pool using refined motion embeddings $\widetilde{\mathbf z}_i^{\mathrm{mot}}$ that combine the anchor and predicted-offset encodings:
\begin{equation}
\begin{aligned}
\mathbf F_i^0&=F_{\mathrm{FAR}}\!\left(\mathbf z^{\mathrm{vis}},\widetilde{\mathbf z}_i^{\mathrm{mot}},\mathbf Q_{\mathrm{fut}};\boldsymbol\eta_0\right),\\
s_i&=R_{\mathrm{FAR}}\!\left(\mathbf F_i^0,\widetilde{\mathbf z}_i^{\mathrm{mot}};\boldsymbol\eta_0\right).
\end{aligned}
\label{eq:s-far-interface}
\end{equation}
Here, $\mathbf Q_{\mathrm{fut}}$ denotes the future-scene query, $\boldsymbol\eta_0$ contains frozen FAR parameters, and $R_{\mathrm{FAR}}$ comprises the remaining decoder and scoring layers. Candidate order is preserved across features, trajectories, and scores. FAR's inherited pretraining is detailed in Appendix~\ref{app:far}.

\subsection{Physical execution descriptor}\label{app:descriptor}

The PRS descriptor is $\mathbf u_E=(\mathbf X_E,\mathbf v_E)$, where $\mathbf X_E\in\mathbb R^{16\times31}$ and $\mathbf v_E\in\mathbb R^{10}$. We sample 16 temporal locations from the reference rollout, including its initial and terminal states. The 31 pointwise channels describe nominal motion, executed motion, tracking errors, nominal and executed corner clearances, boundary directions and distances, time, and validity masks. The ten static features encode ego speed, acceleration and steering rate, vehicle dimensions, controller response parameters, tracking horizon, and initial clearance.

A shared pointwise MLP encodes the temporal features. Its outputs are combined with a static-feature encoding to predict two repair parameters. PRS is pretrained on clearance-aware demonstrations and remains frozen when the future projection and residual predictor are trained. The planner and FAR also remain frozen.

\input{appendix/00a_fcrs_architecture}

\subsection{Optimization}\label{app:training}

We use PyTorch on a single NVIDIA GeForce RTX 5090. AdamW trains the future projection and residual predictor with the settings in Table~\ref{tab:s-training}. The loss is Eq.~\eqref{eq:14} of the main paper. Its parameter scale is $s_\theta=0.5$, the mean Huber loss has unit transition, and trajectory imitation sums mean squared position and wrapped heading errors normalized by 0.75 m and 0.20 rad, respectively. PRS pretraining uses the same loss terms with its physics-only prediction.

\par\addvspace{4pt}
\noindent\begin{minipage}{\columnwidth}
\captionsetup{type=table,skip=4pt}
\centering
\caption{\textbf{Optimization settings for FCRS training.}}
\label{tab:s-training}
\small
\begin{tabular}{@{}lr@{}}
\toprule
Setting & Value \\
\midrule
Updates & 750 \\
Learning rate & $3\times10^{-4}$ \\
Weight decay & $10^{-4}$ \\
Batch size & 64 \\
Repair / already-feasible examples & 48 / 16 \\
Action / trajectory / preservation weights & 1 / 0.25 / 0.10 \\
\bottomrule
\end{tabular}
\end{minipage}
\par\addvspace{4pt}

\subsection{Execution and CAST search}\label{app:execution}

An LQR tracker and kinematic bicycle model simulate four seconds of execution with a 0.1-second step: $N_e=40$ intervals and 41 states including the current pose. Acceleration and steering response constants are 0.20 s and 0.05 s; the tracking horizon is ten steps. At each state, signed road clearance is the minimum over the vehicle's four footprint corners. EAS intervenes at at least 5 cm of executed penetration and prefers the highest-scoring candidate with nonnegative clearance. Repair additionally requires valid execution geometry and an empty road-compliant candidate set. SRV checks at least 2 cm clearance, a maximum absolute wrapped waypoint-heading difference of 0.20 rad from the reference, finite states, and the displacement bound enforced by the deformation.

Starting from zero deformation, CAST minimizes mean squared clearance shortfall over future execution states,
\begin{equation}
\begin{aligned}
J(\boldsymbol\theta)&=\frac{1}{N_e}\sum_{k=1}^{N_e}
\left[\frac{0.10-d_k(\boldsymbol\theta)}{0.25}\right]_+^2\\
&\quad+0.05\,J_{\mathrm{disp}}(\boldsymbol\theta)
+0.01\,J_{\mathrm{smooth}}(\boldsymbol\theta),
\end{aligned}
\label{eq:s-search-objective}
\end{equation}
Here, $d_k(\boldsymbol\theta)$ is signed executed footprint clearance and $[x]_+=\max(x,0)$. $J_{\mathrm{disp}}$ is mean squared lateral displacement and $J_{\mathrm{smooth}}$ is mean squared second difference of lateral displacement, both normalized by 0.75 m. Central finite differences use probes of 0.10, with four parameter probes per update followed by one normalized descent proposal of length 0.35. The reference rollout is reused, limiting two updates to ten additional physical rollouts.

Teacher and student use the same deformation, controller, vehicle model, and acceptance rules. Among qualifying search proposals, CAST maximizes the minimum executed clearance over all states, including the initial pose. Solutions within $10^{-4}$ m of the maximum are ordered by smaller RMS lateral deformation and then stable proposal order. The student instead predicts one proposal followed by one verification rollout. All repair costs exclude shared candidate planning and EAS; cached latency measures the single-thread CPU correction stage.

%% file: appendix/00a_fcrs_architecture.tex
\subsection{Future-conditioned repair student}\label{app:fcrs}

\begin{figure}[!t]
\centering
\includegraphics[width=0.98\columnwidth]{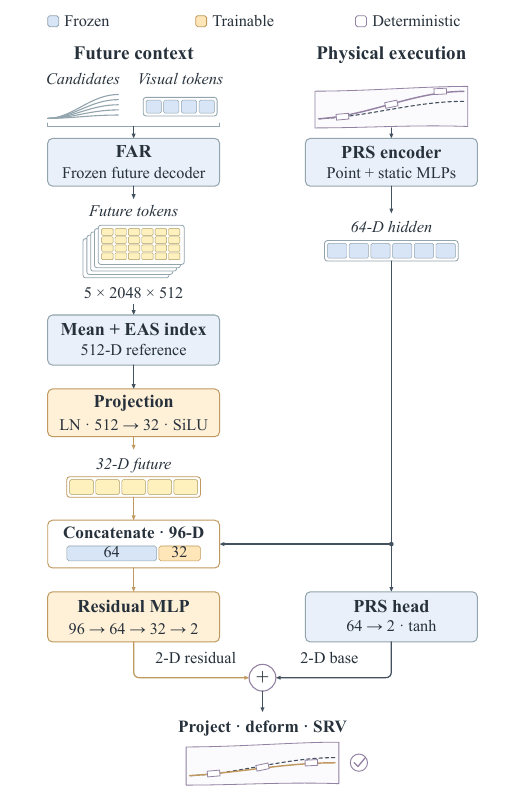}
\caption{\textbf{FCRS architecture.} The future feature of the EAS reference is projected to 32 dimensions and concatenated with the frozen PRS's 64-dimensional hidden representation. A two-dimensional residual is added to the original repair parameters before deformation and verification. Blue marks frozen paths, warm gold the trainable branch, and purple deterministic steps.}
\label{fig:s-fcrs}
\end{figure}

\textbf{Future feature.} Figure~\ref{fig:s-fcrs} shows the future-feature and residual branches. For each of the five retained candidates, we extract $\mathbf F_i^0\in\mathbb R^{2048\times512}$ from the frozen FAR Future-scene Decoder before its $512\!\rightarrow\!16$ projection. We average the 2048 feature-grid tokens within each candidate, select the EAS reference index, and project its vector:
\begin{equation}
\begin{aligned}
\mathbf f_i&=\frac{1}{2048}\sum_{j=1}^{2048}\mathbf F_i^0[j],\\
\mathbf z_F&=\operatorname{SiLU}\!\left(W_F\operatorname{LN}(\mathbf f_{i_{\mathrm{EAS}}})+\mathbf b_F\right).
\end{aligned}
\label{eq:s-future-projection}
\end{equation}
Here, $\mathbf f_i\in\mathbb R^{512}$ and $\mathbf z_F\in\mathbb R^{32}$. LayerNorm has no affine parameters. Pooling is performed separately for each candidate and requires no additional FAR evaluation.

\textbf{Hidden-feature fusion.} The frozen PRS uses pointwise and static MLPs of widths $31\!\rightarrow\!64\!\rightarrow\!64$ and $10\!\rightarrow\!64\!\rightarrow\!64$, respectively, with SiLU activations. Flattening the 16 point encodings and concatenating the static encoding gives 1088 dimensions. A $1088\!\rightarrow\!128\!\rightarrow\!64$ trunk with SiLU produces $\mathbf z_g$; its original linear head and $\tanh$ produce $\boldsymbol\theta_{\mathrm{phys}}$. The added MLP $H_{\boldsymbol\phi}$ takes $[\mathbf z_g;\mathbf z_F]\in\mathbb R^{96}$ and uses widths $96\!\rightarrow\!64\!\rightarrow\!32\!\rightarrow\!2$, with SiLU after its first two layers. Its residual is added as in Eq.~\eqref{eq:10}. The residual and sum have no additional output nonlinearity. Projection and deformation follow Eq.~\eqref{eq:11}, followed by SRV.

\textbf{Initialization and size.} The residual MLP's final weights and biases are zero-initialized, preserving PRS at initialization. Only the future projection and residual MLP are trained, with 16,416 and 8,354 parameters. Together with the 158,850 frozen PRS parameters, the student contains 183,620 parameters, of which 24,770 are trainable. These counts exclude the inherited planner and FAR.

%% file: appendix/02_far_origin.tex
\section{FAR pretraining and its relation to the world model}\label{app:far}

\begin{figure*}[!t]
\centering
\includegraphics[width=\textwidth]{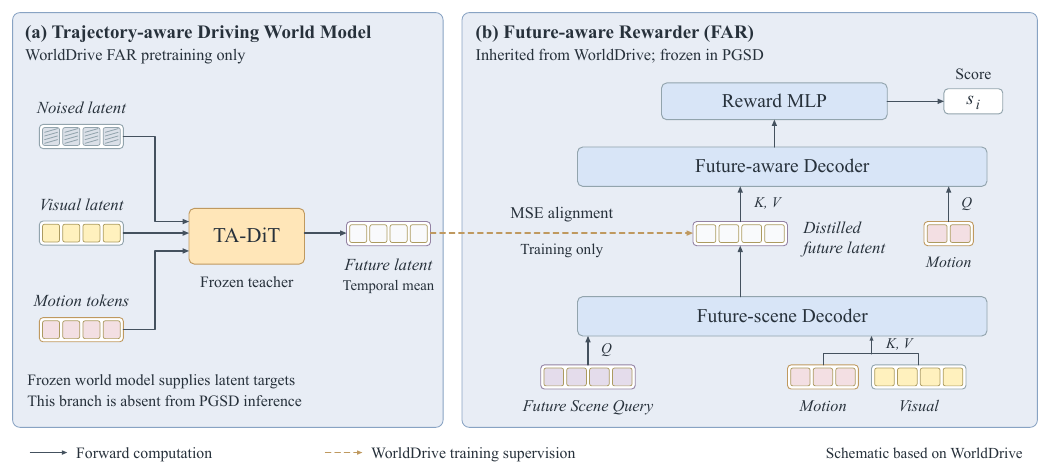}
\caption{\textbf{Origin of the inherited FAR.} In WorldDrive, a frozen trajectory-aware world model supplies future latent targets for FAR pretraining. The Future-scene Decoder predicts candidate-conditioned latent features aligned to these targets through MSE supervision. The Future-aware Decoder and reward MLP score candidates using the distilled representation. Dashed warm-gold arrows denote WorldDrive training supervision. PGSD uses the resulting pretrained FAR and keeps it frozen.}
\label{fig:s1}
\end{figure*}

PGSD inherits WorldDrive's planning backbone and pretrained Future-aware Rewarder (FAR) \cite{gui2026worlddrive}. Figure~\ref{fig:s1} summarizes how world-model future representations are distilled into the frozen rewarder.

During WorldDrive pretraining, learnable future-scene queries attend to historical visual and candidate-motion tokens through the Future-scene Decoder. Its output is aligned with latent targets generated by the frozen world model. These targets are averaged over future latent frames before alignment. The Future-aware Decoder queries the distilled representation using the candidate motion embedding, and the reward MLP produces a candidate score. The schematic omits feature projection and residual pooling.

PGSD uses the pretrained WorldDrive stage-2 planner and freezes FAR during both repair-student training and inference. Scores and candidate-aligned features are reused for selection and repair. The connection to the world model is inherited through FAR's pretrained representations; PGSD requires no diffusion sampling at inference. CAST supervision for the repair student is separate from the FAR pretraining shown here.

%% file: appendix/03a_clearance_distribution_figure.tex
\begin{figure*}[!t]
\centering
\includegraphics[width=0.78\textwidth]{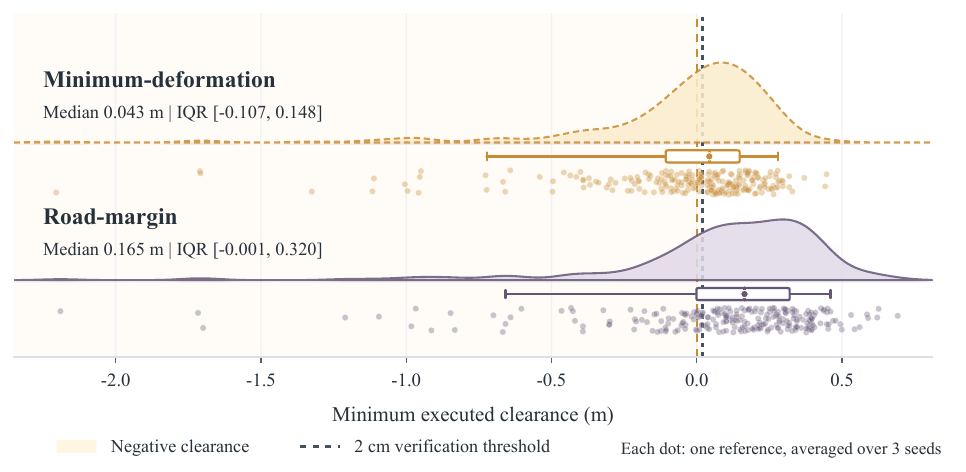}
\caption{\textbf{Execution-clearance distributions.} Each dot is one reference trajectory's mean minimum executed clearance across three seeds. Half-densities share a Gaussian-kernel bandwidth and vertical scale. Boxes show median and IQR; whiskers show the 5th and 95th percentiles. The full linear range retains negative tails. Pale gold denotes negative clearance; the dashed line marks 0.02 m.}
\label{fig:s2}
\end{figure*}

%% file: appendix/01_evaluation.tex
\section{Evaluation protocols and metrics}\label{app:metrics}

NAVSIM v1 \cite{dauner2024navsim} is built on the nuPlan-derived OpenScene dataset and provides approximately 103k training scenarios. We evaluate all 12,146 navtest scenarios with four-second non-reactive rollouts. NAVSIM v2 \cite{cao2025pseudosimulation} uses reactive traffic; navhard\_two\_stage comprises 5,912 frames in 225 groups. CAST demonstrations are generated on navtrain, with training and development separated by recording session and each reference kept with all its search proposals.

For unit-scale subscores, the Predictive Driver Model Score and its extended form are defined as
\begin{equation}
\mathrm{PDMS}=\mathrm{NC}\cdot\mathrm{DAC}\cdot
\frac{5\mathrm{EP}+5\mathrm{TTC}+2\mathrm{C}}{12},
\label{eq:s-pdms}
\end{equation}
\begin{equation}
\begin{aligned}
\mathrm{EPDMS}&=\frac{\mathrm{NC}\cdot\mathrm{DAC}\cdot\mathrm{DDC}\cdot\mathrm{TLC}}{16}\\
&\quad\cdot\bigl(5\mathrm{EP}+5\mathrm{TTC}+2\mathrm{LK}\\
&\qquad+2\mathrm{HC}+2\mathrm{EC}\bigr).
\end{aligned}
\label{eq:s-epdms}
\end{equation}
NC, DAC, TTC, C, and EP denote no-at-fault collisions, drivable-area compliance, time-to-collision, comfort, and ego progress. EPDMS also includes driving-direction compliance (DDC), traffic-light compliance (TLC), lane keeping (LK), history comfort (HC), and extended comfort (EC). EC measures consistency between plans from successive observations. EPDMS replaces a method subscore with one when the corresponding human subscore is zero.

In NAVSIM v2, Stage 1 evaluates plans from real observations; Stage 2 uses pre-rendered observations at perturbed vehicle states. The aggregate score is
\begin{equation}
s_{\mathrm{agg}}=s_1\sum_j\bar w_j s_{2,j},
\qquad\sum_j\bar w_j=1,
\label{eq:s-two-stage}
\end{equation}
where $s_1$ and $s_{2,j}$ are the first- and second-stage scores. Normalized proximity weights $\bar w_j$ reflect the distance from each second-stage starting state to the first-stage rollout endpoint. We report benchmark scores on the 0--100 scale.

Repair qualification requires the displacement and heading limits of the main paper and at least 2 cm executed road clearance. Rates over all predefined repair examples and over teacher-feasible subsets are distinct. Rejected proposals fall back to the EAS reference, which is not counted as successful repair. Collision and TTC outcomes are evaluated by the native planning metrics. Planning scores cover all test scenarios; Appendices~\ref{app:clearance} and~\ref{app:rejections} analyze the matched physics-only development study.

%% file: appendix/03_clearance_statistics.tex
\section{Execution clearance under clearance-aware supervision}\label{app:clearance}

\textbf{Matched study and statistical units.} We compare physics-only repair students using full execution feedback and either minimum-deformation targets ($G_{\mathrm{min}}$) or road-margin targets ($G_{\mathrm{clear}}$). The policies share the architecture, search-proposal library, training budget, development set, and three paired seeds. All 241 execution-infeasible development reference trajectories are retained, drawn from 143 scenes in 32 recording sessions. Each policy produces 723 proposals. Both accepted and rejected proposals are included, even for references without a qualifying teacher target.

For policy $p$, let $m_{r,s}^{p}$ be the minimum signed executed corner clearance for reference $r$ and seed $s$. We average within each reference and compute the paired gain:
\begin{equation}
\begin{aligned}
\bar m_r^{p}&=\frac{1}{3}\sum_{s=0}^{2}m_{r,s}^{p},\\
\Delta_r&=\bar m_r^{G_{\mathrm{clear}}}-\bar m_r^{G_{\mathrm{min}}}.
\end{aligned}
\label{eq:s-paired-gain}
\end{equation}
We weight references equally. To account for within-scene and within-session dependence, we compute 10,000 paired bootstrap draws over recording sessions, retaining all references and both policies from each sampled session. The intervals condition on the three paired seeds; seed variability is reported separately. All references, including those without qualifying teacher targets, enter this analysis.

\textbf{Distributional shift.} Figure~\ref{fig:s2} shows that road-margin supervision increases the median reference-averaged clearance from 4.32 to 16.45 cm. The IQR shifts from $[-10.67,14.77]$ cm to $[-0.09,31.98]$ cm (Table~\ref{tab:s-clearance}). Both policies retain negative tails: their minimum reference-averaged values are $-2.205$ and $-2.190$ m. These cases still require verification and fallback.

\par\noindent\begin{minipage}{\columnwidth}
\captionsetup{type=table,skip=4pt}
\centering
\caption{\textbf{Matched demonstration-selection results.} Acceptance is mean $\pm$ sample SD across three seeds.}
\label{tab:s-clearance}
\small
\setlength{\tabcolsep}{3pt}
\begin{tabular}{@{}lcc@{}}
\toprule
Targets & \shortstack{Median [IQR]\\clearance (cm)} & \shortstack{Acceptance\\(\%)} \\
\midrule
Minimum-deformation & \shortstack{4.32\\$[-10.67,14.77]$} & $53.0\pm1.0$ \\
Road-margin & \shortstack{16.45\\$[-0.09,31.98]$} & $70.1\pm1.1$ \\
\bottomrule
\end{tabular}
\end{minipage}\par\addvspace{6pt}

\textbf{Paired gains and regressions.} Clearance increases for 228 of 241 references (94.6\%), decreases for 11, and is unchanged for two. The median paired gain is 14.20 cm (95\% session-bootstrap interval: $[11.40,16.72]$ cm); the mean is 13.63 cm ($[11.54,15.28]$ cm). The proportion of improved references has a 95\% interval of $[90.1\%,97.9\%]$. Negative paired gains appear to the left of zero in Figure~\ref{fig:s3}. The paired median summarizes reference-level differences and need not equal the difference of marginal medians.

%% file: appendix/03b_paired_clearance_figure.tex
\noindent\begin{minipage}{\textwidth}
\captionsetup{type=figure,skip=4pt}
\centering
\includegraphics[width=0.84\textwidth]{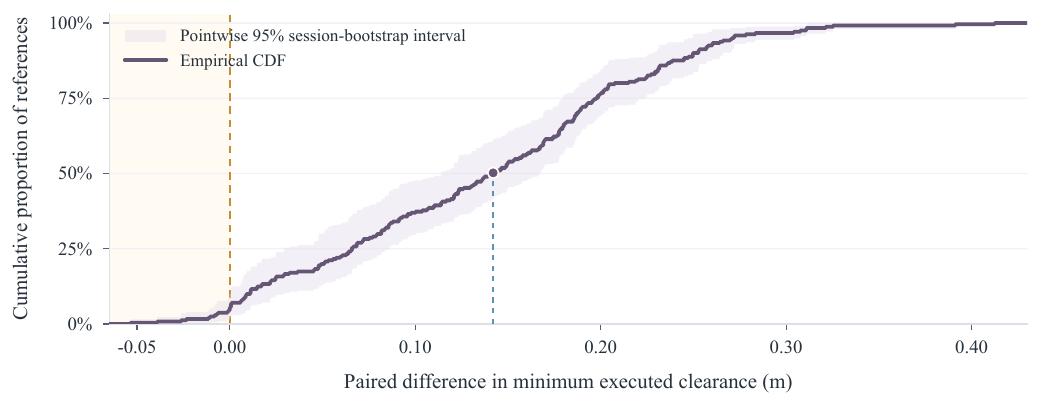}
\caption{\textbf{Empirical CDF of paired execution-clearance gains.} The horizontal coordinate is $\Delta_r$ for the same reference trajectory under both policies after averaging its three seeds. Positive gains favor road-margin supervision. The shading is a pointwise 95\% session-bootstrap interval, not a simultaneous band. The curve is the exact empirical step function, and the highlighted point is the median paired gain.}
\label{fig:s3}
\end{minipage}\par

%% file: appendix/04_rejections.tex
\section{Acceptance and recorded rejection reasons}\label{app:rejections}

\noindent\begin{minipage}{\textwidth}
\captionsetup{type=figure,skip=4pt}
\centering
\includegraphics[width=0.76\textwidth]{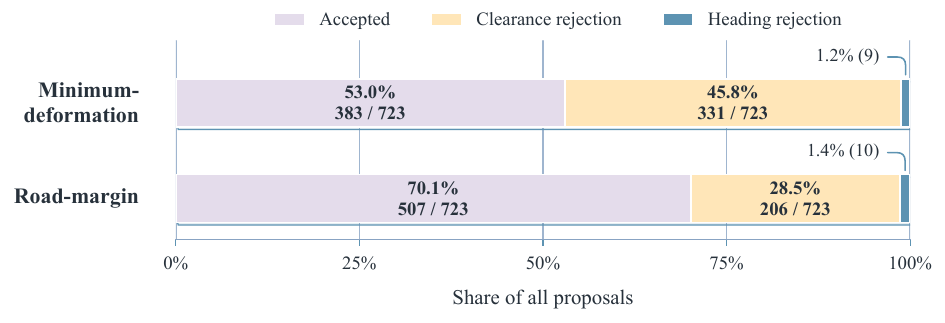}
\caption{\textbf{Acceptance and recorded rejection reasons.} Each bar includes all 723 proposals for one policy (241 reference trajectories $\times$ three seeds), without treating these as independent scenes. Acceptance requires the full verification rule. Fallback is not successful repair. Rejection categories follow a fixed priority: heading before clearance. Heading-rejected proposals may also fail clearance checks.}
\label{fig:s4}
\end{minipage}\par\addvspace{5pt}

Road-margin supervision increases accepted proposals from 383/723 (53.0\%) to 507/723 (70.1\%). Across paired seeds, acceptance is $53.0\pm1.0\%$ and $70.1\pm1.1\%$ (mean $\pm$ sample SD). The gain, with equal weight per reference, is 17.15 percentage points, with paired session-bootstrap 95\% interval $[11.88,21.79]$ points.

The gain is concentrated in recorded clearance rejections, which decrease from 331/723 (45.8\%) to 206/723 (28.5\%). Heading rejections remain uncommon, changing from 9/723 (1.2\%) to 10/723 (1.4\%). This pattern is consistent with clearance-aware targets providing a geometric buffer for imitation errors. The counts describe the verifier's recorded outcomes, not disjoint physical failure mechanisms or collision risk. Together with the clearance distributions, they support clearance-aware demonstration selection in the matched physics-only study.

%% file: appendix/05_hugsim.tex
\section{Qualitative zero-shot planning on HUGSIM}\label{app:hugsim}

\begin{figure}[!ht]
\centering
\includegraphics[width=\textwidth]{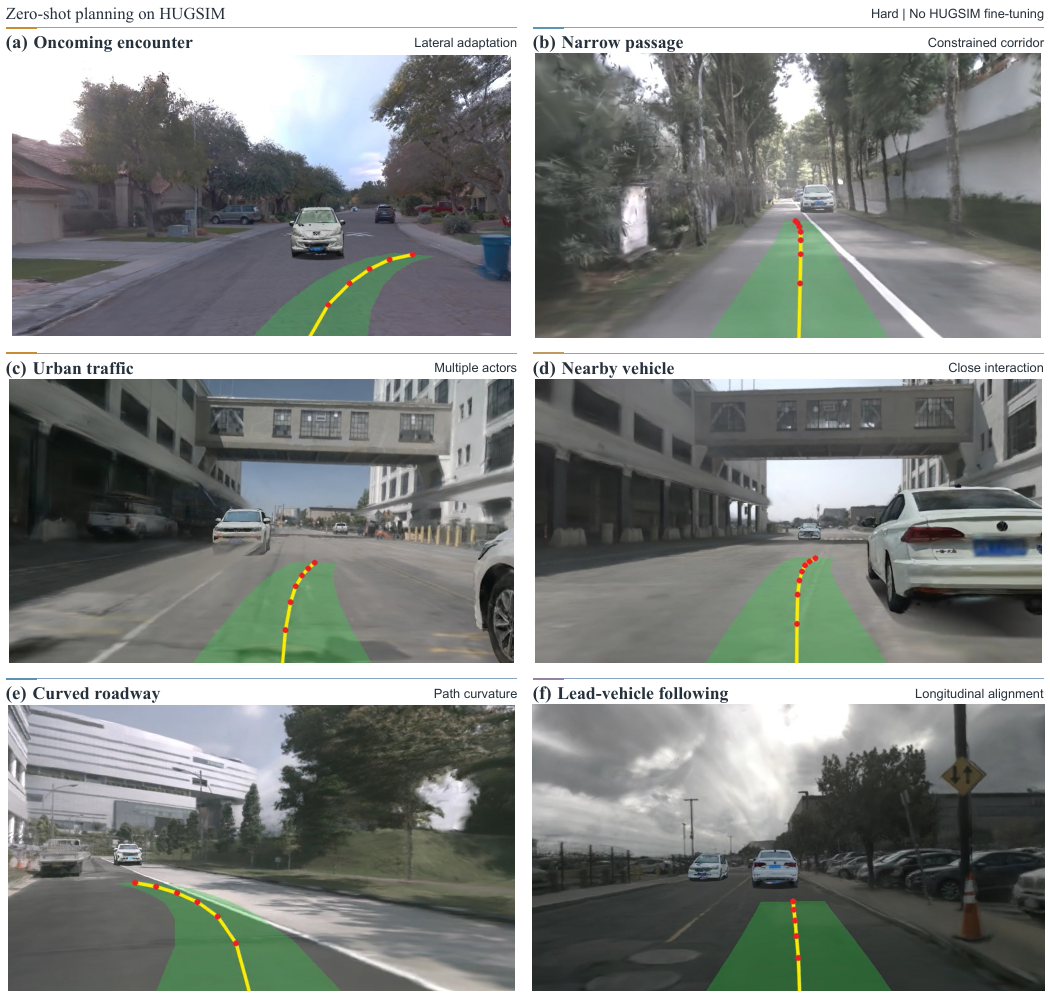}
\caption{\textbf{Qualitative zero-shot planning examples on HUGSIM Hard.} Selected front-camera views without HUGSIM-specific training or fine-tuning show (a) an oncoming encounter, (b) a narrow passage, (c) urban traffic, (d) a nearby vehicle, (e) a curved roadway, and (f) lead-vehicle following. Yellow lines and red dots denote paths and waypoints. Each panel shows one planning snapshot.}
\label{fig:s5}
\end{figure}

We present qualitative zero-shot planning examples on HUGSIM Hard \cite{zhou2026hugsim}, obtained without simulator-specific training or fine-tuning. Figure~\ref{fig:s5} shows plans following narrow and curved corridors, adapting lateral placement around nearby vehicles, and aligning with a lead vehicle. These snapshots illustrate scene-conditioned planning geometry beyond NAVSIM.

